\documentclass{article}

     \usepackage[final]{neurips_2018}

\usepackage[utf8]{inputenc} 
\usepackage[T1]{fontenc}    
\usepackage{hyperref}       
\usepackage{url}            
\usepackage{booktabs}       
\usepackage{amsfonts}       
\usepackage{nicefrac}       
\usepackage{microtype}      
\usepackage{graphicx}
\usepackage{amsmath}

\title{Over-smoothing Effect of Graph Convolutional
Networks}

\author{%
  Fang Sun\\
  EECS, Peking University\\
  \texttt{fts@pku.edu.cn} \\
}

\begin{document}

\maketitle

\begin{abstract}
  Over-smoothing is a severe problem which limits the depth of Graph Convolutional Networks. This article gives a comprehensive analysis of the mechanism behind Graph Convolutional Networks and the over-smoothing effect. The article proposes an upper bound for the occurrence of over-smoothing, which offers insight into the key factors behind over-smoothing. The results presented in this article successfully explain the feasibility of several algorithms that alleviate over-smoothing.
\end{abstract}

\section{Introduction}

Graph data are ubiquitous: from social networks like Weibo and Twitter, to citation graphs connecting knowledge production in academia, they provide a natural and flexible way of presenting the information we generate everyday. Graph Convolutional Network (GCN) [Kipf \& Welling, 2017] is a successful attempt to generalize the powerful convolutional networks (CNNs) in coping with Euclidean data to modeling graph structured data. GCN is simple and elegant, out-performing previous works by a large margin on semi-supervised classification tasks.

Recently, I performed an experiment on vanilla GCN. I stacked up the convolutional layers in GCN from 2 to 3, 4, 5, 6, and tested their power on Cora citation network data-set. The result (\textbf{Figure 1}) was astounding: although GCN falls in the category of  'deep' learning, its power quickly diminishes as its layers stack up to merely 6. Several possible explanations for this phenomenon were quickly ruled out: Not over-fitting, because training accuracy and testing accuracy degraded synchronously. Not vanishing gradient, since a 6-layer network is too shallow for such an effect to occur. 

\begin{figure}[h]
  \centering
  \includegraphics[scale=0.5]{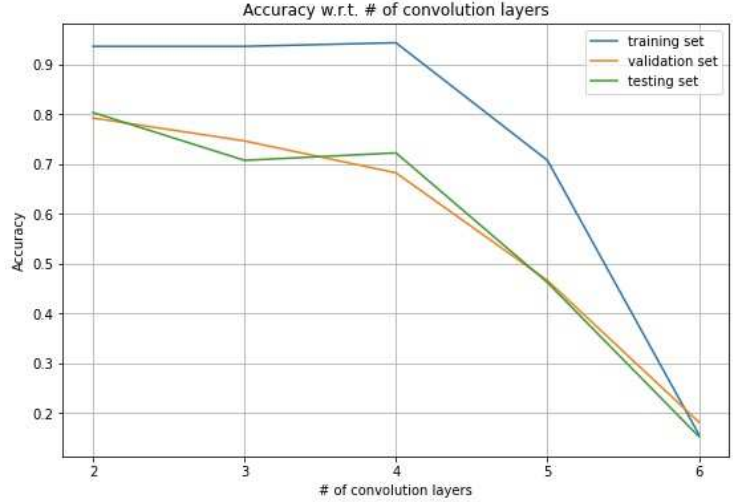}
  \caption{Accuracy of GCN on Cora w.r.t. \# of convolution layers.}
\end{figure}

In fact, this effect is unique to GCNs, called over-smoothing. [Li et al., 2018] proposes the concept, pointing out that the intrinsic smoothing nature of graph Laplacians limits the deepening of GCNs. Much work has been done in the last 2 years to address the over-smoothing problem.

These are the 3 main objectives I am trying to establish in this article:

\begin{itemize}
    \item Giving mathematical formulations of 'smoothness';
    \item Analyzing the mechanisms for over-smoothing and the scenarios under which over-smoothing could happen;
    \item Explaining different methods for alleviating over-smoothing.
\end{itemize}

\section{Preliminaries for Graph Convolution}

\subsection{Basic Architecture of GCN}

\paragraph{Graph Laplacian}
GCN is essentially a neighborhood-augmented MLP. Inspired by signal processing, GCN uses the Laplacian matrix to aggregate the neighborhood information. For graph \(G\), Laplacian matrix \(\boldsymbol L:=\boldsymbol D- \boldsymbol A\). \(\boldsymbol D\) is the degree matrix of \(G\), and \(\boldsymbol A\) is the adjacency matrix of \(G\).

The spectral convolution of GCN is presented as:
\[\boldsymbol H^{(l+1)}=\sigma\left(\boldsymbol{\tilde{D}}^{-\frac{1}{2}}  \boldsymbol{\tilde{A}} \boldsymbol {\tilde{D}}^{-\frac{1}{2}} \boldsymbol H^{(l)} \boldsymbol \Theta^{(l)}\right).\]
\(\boldsymbol H^{(l)}, \boldsymbol H^{(l+1)}\) are the outputs of the previous/present layer, \(\boldsymbol \Theta^{(l)}\) is the tune-able convolution kernel, \(\sigma\) is the non-linear activation function (ReLU).

\paragraph{Derivation of GCN Model}
The GCN model is derived via 4 steps of approximation: 
\begin{align*}
g_{\theta} \star \boldsymbol x
&=g_{\theta} \boldsymbol U^{\top} \boldsymbol x &(\text{spectral convolution})\\
&\approx \sum_{k=0}^{K} \theta_{k}^{\prime} T_{k}(\boldsymbol {\tilde{L}}) \boldsymbol x
&(k\text{-th order Chebyshev apprm.})\\
&\approx \theta_{0}^{\prime} \boldsymbol x+\theta_{1}^{\prime}\left(\boldsymbol L-\boldsymbol I_{N}\right)\boldsymbol  x
& (k=1, \text{considering only 1st order neighbor})\\
&\approx \theta\left(\boldsymbol I_{N}+\boldsymbol D^{-\frac{1}{2}} \boldsymbol A \boldsymbol D^{-\frac{1}{2}}\right)\boldsymbol  x\\
&\approx \theta\left(\boldsymbol {\tilde{D}}^{-\frac{1}{2}} \boldsymbol {\tilde{A}} \boldsymbol {\tilde{D}}^{-\frac{1}{2}}\right) \boldsymbol x
&\text{(the re-normalization trick)}
\end{align*}
The re-normalization trick is \(\boldsymbol I_{N}+\boldsymbol D^{-\frac{1}{2}} \boldsymbol A \boldsymbol D^{-\frac{1}{2}} \rightarrow \boldsymbol {\tilde{D}}^{-\frac{1}{2}} \boldsymbol{\tilde{A}} \boldsymbol {\tilde{D}}^{-\frac{1}{2}}\). \(\boldsymbol{\tilde{D}}^{-\frac{1}{2}} \boldsymbol{\tilde{A}} \boldsymbol{\tilde{D}}^{-\frac{1}{2}}\) can be viewed as the normalized Laplacian matrix of graph \(\widetilde{G}\), i.e. \(G\) with self-loop. The implication behind taking \(1\)st order Chebyshev approximation is that, in each layer of convolution, the model only considers the \(1\)st order neighbor of each node. Nevertheless, higher orders of neighbor information can be aggregated via stacking more convolution layers.

\subsection{Laplacian Smoothing is the Key Power of GCN}

[Li et al., 2018] proposes that Laplacian smoothing is central to GCN's power in classification tasks. The layer-wise propagation rule of the simplest fully-connected networks (FCNs) is
\[\boldsymbol H^{(l+1)}=\sigma(\boldsymbol H^{(l)}\boldsymbol \Theta^{(l)}).\]

We observe that the sole difference between GCN and FCN is the \textbf{normalized Laplacian matrix} \(\boldsymbol S=\boldsymbol {\tilde{D}}^{-\frac{1}{2}} \boldsymbol {\tilde{A}} \boldsymbol {\tilde{D}}^{-\frac{1}{2}}\). By comparison, even a 1-layer GCN can out-perform a 1-layer FCN by a large margin. This is because Laplacian smoothing makes the output features of nodes in the same cluster more similar and eases the classification task.

The aggregating abilities of Laplacian smoothing is further demonstrated by Simple Graph Convolution (SGC) [Wu et al., 2019]:
\[\boldsymbol{\hat{{Y}}}_{\mathrm{SGC}}=\operatorname{softmax}\left({\boldsymbol S}^{K} {\boldsymbol X} {\boldsymbol \Theta}\right),\]
where \(K\) is the number of Laplacian matrices stacked. SGC shows that even if we remove the redundant ReLU (non-linearity) and MLP layers between aggregators, the multi-layer Laplacian smoothing yields the same degree of accuracy with GCN.

Yet, by applying Laplacian smoothing many times, the feature of nodes in the same connected component will converge to the same value and thus become indistinguishable. As is shown in \textbf{Figure 2}, while the two types of points are well-separable under the 2-layer scenario, they all become squashed up in the 5-layer GCN. 

\begin{figure}[h]
  \centering
  \includegraphics[scale=0.28]{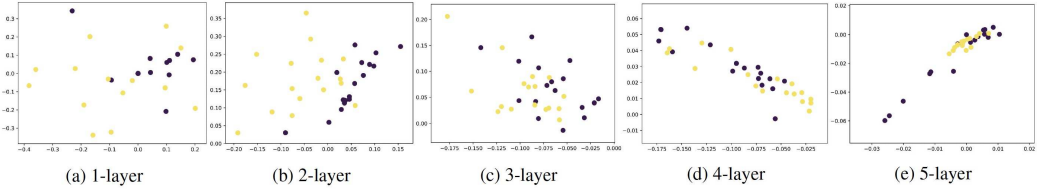}
  \caption{Vertex embeddings of Zachary’s karate club network with GCNs of 1,2,3,4,5 layers.}
\end{figure}

Thus, we give the natural definition of \textbf{over-smoothing}.
\paragraph{Definition 1}
\textbf{Over-smooothing} is the effect that node features become indistinguishable after multiple rounds  of Laplacian smoothing.

\section{Deeper Insight into Over-smoothing via Mathematical Formulation}

\subsection{Spectral Analysis of GCN}

Recent works addressing the over-smoothing issue tend to regard GCN as low-pass filtering [NT \& Maehara, 2019], inspired by signal processing. The spectral analysis on GCN has yielded some qualitative insight into the issue.
\paragraph{Theorem 1}
Given a connected graph \(G\), for the normalized
Laplacian \(\boldsymbol S\),
\[\lim_{k\to\infty}{\boldsymbol S}^k=\boldsymbol \Pi,\]
where \(\boldsymbol {\Pi}=\Phi\left(\boldsymbol{\widetilde{{D}}}^{\frac{1}{2}}\boldsymbol {e}^{\top}\right)\left(\Phi\left(\boldsymbol {\widetilde{{D}}}^{\frac{1}{2}} \boldsymbol{e}^{\top}\right)\right)^{\top}\) , \(\Phi(\mathbf{x}) = \frac{\mathbf{x}}{||\mathbf{x}||}\).

\textbf{Proof:} Because \(\boldsymbol  S\) is symmetric, we orthogonally diagonalize \(\boldsymbol S=\boldsymbol Q\boldsymbol \Lambda \boldsymbol Q^{\top}\). Thus,
\[\boldsymbol S^{k}=\boldsymbol Q \boldsymbol \Lambda \boldsymbol Q^{\top} \cdots \boldsymbol Q \boldsymbol \Lambda \boldsymbol Q^{\top}=\boldsymbol Q \boldsymbol \Lambda^{k} \boldsymbol Q^{\top}=\sum_{i=1}^{k} \lambda_{i}^{n} \boldsymbol v_{i} \boldsymbol v_{i}^{\top},\]
where \(\boldsymbol v_i\) is the normalized eigenvector of \(\lambda_i\). Laplacian \(\boldsymbol S\) always has an eigenvalue \(1\) with unique associated eigenvector \(\boldsymbol {\widetilde{{D}}}^{\frac{1}{2}} \boldsymbol{e}^{\top}\), and all other eigenvalues \(\lambda\) satisfy \(|\lambda| < 1\). Thus, as \(k\to \infty\), \(\boldsymbol S^k \to \Phi\left(\boldsymbol {\widetilde{{D}}}^{\frac{1}{2}}\boldsymbol {e}^{\top}\right)\left(\Phi\left(\boldsymbol {\widetilde{{D}}}^{\frac{1}{2}} \boldsymbol{e}^{\top}\right)\right)^{\top}=\boldsymbol \Pi\).

\textbf{Theorem 1} demonstrates that over-smoothing is inevitable in
very deep models, where \(\boldsymbol{S}^k\) converges to \(\boldsymbol{\Pi}\). In this scenario, only the degree information of graph \(G\) is retained.

Another important thing to consider is the convergence rate. From the above deduction, the convergence rate is associated with the largest eigenvalue of \(\boldsymbol S\) other than \(1\). If we view from another angle and look at how the features of each node \(\boldsymbol v_i\) in GCN is aggregated with its local neighbors:
\[{\mathbf{h}}_{i}^{(k)} \leftarrow \frac{1}{d_{i}+1} \mathbf{h}_{i}^{(k-1)}+\sum_{i=1}^{N} \frac{a_{i j}}{\sqrt{\left(d_{i}+1\right)\left(d_{j}+1\right)}} \mathbf{h}_{j}^{(k-1)}.\]
The above propagation suggests that the higher the node degree \(d_i\) is, the quicker feature \(\boldsymbol h_i\) would converge. Thus we have the following claim:
\paragraph{Claim 1}
Nodes with higher degree \(d_i\) are more likely to suffer from over-smoothing.

\subsection{Quantifying Smoothness: a Topological View}

In addressing the over-smoothing effect, many papers have proposed their own metric for smoothness, either to quantify and prove their
hypothesis, or to validate the effectiveness of their method. JKNet
{[}Xu et al., 2018{]} defined Influence Score to measure the sensitivity of node \(x\) to node \(y\), and uses the Influence Distribution to capture the relative influences of all other nodes. PairNorm {[}Zhao \& Akoglu, 2020{]} focuses on node-wise smoothing and feature-wise smoothing, and defined two metrics for smoothness: rol-diff and col-diff. {[}Chen et al., 2020{]} proposes Mean Average Distance (MAD). MAD reflects the smoothness of graph representation by calculating the mean of the average distances between nodes. However, these metrics are generally task-specific and incompatible to further theoretical analysis.

By contrast, the metric proposed by {[}Oono \& Suzuki, 2020{]} provides a general framework for measuring smoothness, which solely relies on the topological information of the underlying graph \(G\). Denote the maximum singular value of convolution kernel \(\boldsymbol{\Theta_l}\) by \(s_l\) and set \(s := \sup_{l\in\mathbb{N}_+}s_l\). Denote the distance induced as the Frobenius norm from \(\boldsymbol{X}\) to \(\mathcal{M}\) by \(d_{\mathcal{M}}(\boldsymbol{X}):=\inf _{\boldsymbol{Y} \in \mathcal{M}}\|\boldsymbol{X}-\boldsymbol{Y}\|_{\mathrm{F}}\), where \(\mathcal{M}:=\left\{\boldsymbol{E} \boldsymbol{C} \mid \boldsymbol{C} \in \mathbb{R}^{M \times C}\right\}\), and \(\boldsymbol{E}\) is the eigenspace associated with
\(\lambda_{N-M},\lambda_{N-M+1},\cdots ,\lambda_{N}\). We define the \(\epsilon\)-smoothing metric:
\paragraph{Definition 1}
(\(\epsilon\)-smoothing) If there exists a layer \(L\), such that for any hidden layer \(l\) beyond \(L\), output feature \(\boldsymbol{H}^{(l)}\) has a distance smaller than \(\epsilon\) w.r.t. subspace \(\mathcal{M}\), we call the GCN suffers from \(\epsilon\)-smoothing, i.e.,
\[\exists L, \forall l\ge L, d_{\mathcal{M}}(\boldsymbol{H^{(l)}})< \epsilon.\]

From [Oono \& Suzuki, 2020], we have the following lemma:
\paragraph{Lemma 1}
Let \(\lambda_1\le \cdots\le\lambda_N\) be the eigenvalues of graph Laplacian \(\boldsymbol{S}\), sorted in ascending order. Suppose the multiplicity of the largest eigenvalue \(\lambda_N=1\) is \(M(\le N)\), i.e. \(\lambda_{N-M}<\lambda_{N-M+1}=\cdots =\lambda_{N}=1\). The second largest eigenvalue is defined as
\[\lambda:=\max _{n=1}^{N-M}\left|\lambda_{n}\right|<\left|\lambda_{N}\right|.\]
Then we have
\(\lambda<\lambda_N=1\), and
\[d_{\mathcal{M}}\left(\boldsymbol{H}^{(l)}\right) \leq s_{l} \lambda d_{\mathcal{M}}\left(\boldsymbol{H}^{(l-1)}\right).\]

If all the kernel \(\boldsymbol{\Theta_l}\) have been initialized such that \(s_l < 1\), we have \(s_l\lambda <1\), and the output feature \(\boldsymbol{H}^{(l)}\) \textbf{exponentially} approaches \(\mathcal{M}\) w.r.t. layer depth \(l\).
We derive \textbf{Theorem 2} from \textbf{Lemma 1}:
\paragraph{Theorem 2}
If \(s\lambda < 1\), then \(\epsilon\)-smoothing would happen whenever layer depth \(l\) satisfies
\[l \geq \hat{l}=\left\lceil\frac{\log \frac{\epsilon}{d_{\mathcal{M}}(\boldsymbol{H^{(0)}})}}{\log (s \lambda)}\right\rceil.\]
\textbf{Proof:} From \textbf{Lemma 1}, we have
\begin{align*}
d_{\mathcal{M}}\left(\boldsymbol{H}^{(l)}\right) & \leq s_{l} \lambda d_{\mathcal{M}}\left(\boldsymbol{H}^{(l-1)}\right) \\
& \leq\left(\prod_{i=1}^{l} s_{i}\right) \lambda^{l} d_{\mathcal{M}}(\boldsymbol{H^{(0)}}) \\
& \leq s^{l} \lambda^{l} d_{\mathcal{M}}(\boldsymbol{H^{(0)}})\\
&< \epsilon.
\end{align*}
The inequality is equivalent to
\[l > \frac{\log \frac{\epsilon}{d_{\mathcal{M}}(\boldsymbol{H^{(0)}})}}{\log (s \lambda)}.\]
Taking the ceiling on RHS, we have \(l\ge \hat{l}\).

\section{Main Result: Factors Contributing to Over-smoothing}

\subsection{Upper Bound for the Occurrence of Over-smoothing}

When would the exponential over-smoothing occur? According to \textbf{Theorem 2}, We only need to guarantee that \(s\lambda < 1\). {[}Oono \& Suzuki, 2020{]} has studied the issue on Erdos--Renyi graph \(G_{N, p}\). Here we study the issue in a more generalized setting.
\paragraph{Theorem 3}
For \(N\)-order graph \(G\) with no isolated nodes, denote its largest node degree as \(d_{max}\), and denote its diameter as \(D\). The GCN satisfies the condition of  \textbf{Theorem 2}, i.e. \(s\lambda <1\), providing that
\[s< (1-\frac{4}{NDd_{max}})^{-1}.\]
\textbf{Proof: }We carry out our discussion on graph \(\widetilde{G}\), which adds a self-loop to each node in \(G\).

First, we consider the smallest eigenvalue other than \(0\) of the unnormalized Laplacian \(\boldsymbol{L}=\boldsymbol{\widetilde{D}}-\boldsymbol{\widetilde{A}}\), denoted as \(\lambda(\widetilde{G})\). \(\lambda(\widetilde{G})\) is the famous algebraic connectivity (Fiedler eigenvalue, {[}Fiedler, 1973{]}). According to {[}Mohar, 1991, p. 25{]}, \(\lambda(\widetilde{G})\) is bounded by
\[\lambda(\widetilde{G}) \ge \frac{4}{ND}.\]
Then we consider the relation between \(\lambda(\widetilde{G})\) and \(\lambda\). According to {[}Cavers, 2010{]}, eigenvalues of the normalized Laplacian \(\boldsymbol{S}=\boldsymbol{\tilde{D}^{-\frac{1}{2}} \tilde{A} \tilde{D}^{-\frac{1}{2}}}\) is bounded by their corresponding eigenvalues in \(\boldsymbol{L}\) and \(d_{max}\):
\[\lambda_k(\boldsymbol{S}) \le 1 - \frac {\lambda_k(\boldsymbol{L})}{d_{max}},\]
where \(\lambda_k(\boldsymbol{S})\) is the \(k\)-th largest eigenvalue of \(\boldsymbol{S}\), and \(\lambda_k(\boldsymbol{L})\) is the \(k\)-th smallest eigenvalue of \(\boldsymbol{L}\). Because \(\lambda\) and \(\lambda(\widetilde{G})\) are corresponding eigenvalues, take them into the above inequality to derive
\begin{align*}
\lambda 
&\le 1- \frac{\lambda(\widetilde{G})}{d_{max}}\\
&\le 1-\frac{4}{NDd_{max}}.
\end{align*}
It suffices to show \(s\lambda < 1\) if we set
\[s< (1-\frac{4}{NDd_{max}})^{-1}.\]
Thus, the condition of \textbf{Theorem 2} is satisfied, and exponential over-smoothing could happen in this scenario.

\subsection{What Factors Contribute to Over-smoothing?}

\paragraph{Large and Dense Graphs}
Large and dense graphs suffer from over-smoothing. The conclusion is in line with {[}Oono \& Suzuki, 2020{]}, which formulates the issue on Erdos--Renyi graphs. It also confirms the sensibility of graph sparsification methods for combating over-smoothing, e.g. DropEdge {[}Rong et al., 2020{]}.

\paragraph{Small-World Graphs}
Small-world graphs, with \(D\propto log\ N\), have already achieved relatively high performance on GCNs with only \(2\sim 3\) layers, since these \(2\sim 3\) hops are sufficient to aggregate neighboring information from a large portion of the whole graph. By contrast, tasks like Point Cloud Classification requires deeper convolutions to capture long-range information.

\paragraph{GCN with Residual Connection}
In theory, adding residual connections alone cannot address the over-smoothing issue. If we regard graph convolution as a Markov process {[}Oono \& Suzuki, 2020{]}, the residual connection only leads to a lazy version of the Markov process. The graph Laplacian would still converge, as is shown in \textbf{Theorem 1}. Effective versions of residual connections will be discussed in the next section.

\section{Methods for Alleviating Over-smoothing}

\subsection{Leveraging between Different Convolution Depths}

\paragraph{DAGNN}
{[}Liu et al., 2020{]} This SGC-based {[}Wu et al., 2019{]} work is straight-forward, simple and elegant. With insight from \textbf{Claim 1} that node features are smoothed at different rates w.r.t. node degree, DAGNN stacks up the features output from different convolution depths. By adaptively adjusting these features, DAGNN exploits the advantage of deeper Laplacian convolutions without suffering from performance degradation. The adaptive adjustment process of DAGNN is shown above, where \(\boldsymbol s\) is a trainable projection vector.
\begin{align*}
&\boldsymbol Z = \text{MLP}(\boldsymbol \Theta);\\
&\boldsymbol{H}_l= \boldsymbol S^l \boldsymbol Z, l=1,2,\cdots, k;\\
&\boldsymbol H = \text{stack}(\boldsymbol Z, \boldsymbol H_1, \cdots, \boldsymbol H_k);\\
& \boldsymbol S = \sigma(\boldsymbol{Hs}).
\end{align*}

\subsection{Graph Sparsification}

\paragraph{DropEdge}
{[}Rong et al., 2020{]} As has been discussed in \textbf{Section 4.2}, graph sparsification can slow down the convergence rate of over-smoothing by reducing information passage between layers. DropEdge randomly removes a certain number of edges from the input graph at each training epoch, and can be equipped to many other backbone models. The removal of edges in DropEdge is dynamic and layer-wise:
\[\boldsymbol{H}^{(l+1)}=\sigma\left(\mathfrak{N}\left(\boldsymbol{A} \odot \boldsymbol{Z}^{(l)}\right) \boldsymbol{H}^{(l)} \boldsymbol{\Theta}^{(l)}\right),\]
where \(\boldsymbol{Z}^{(l)}\) is the binary random mask, and \(\mathfrak{N}(\cdot)\) is the normalization operator, i.e. \(\mathfrak{N}(\boldsymbol{A})=\boldsymbol{I}_{N}+\boldsymbol{D}^{-1 / 2} \boldsymbol{A} \boldsymbol{D}^{-1 / 2}.\)

Other 'smarter' ways of dropping edges include Graph DropConnect (GDC) {[}Hasanzadeh et al., 2020{]}, which drops edges both layer-wise and channel-wise. Also, NeuralSparse {[}Zheng et al., 2020{]} uses neural networks to drop out edges.

\subsection{Adding Residual Connections}

\paragraph{GCNII}
 {[}Chen et al., 2020{]} This is the first work that successfully trains deep GCNs on knowledge graphs, with up to \(64\) layers of Laplacian. The propagation rule of GCNII is
\[\boldsymbol{H}^{(\ell+1)}=\sigma\left(\left(\left(1-\alpha_{\ell}\right) {\boldsymbol{S}} \boldsymbol{H}^{(\ell)}+\alpha_{\ell} \boldsymbol{H}^{(0)}\right)\left(\left(1-\beta_{\ell}\right) \boldsymbol{I}_{n}+\beta_{\ell} \boldsymbol{\Theta}^{(\ell)}\right)\right)\]
The identity mapping (\(\left(1-\beta_{\ell}\right) \boldsymbol{I}_{n}+\beta_{\ell} \boldsymbol{\Theta}^{(\ell)}\)) resembles that of ResNet, yet the initial residual connection (\(\left(1-\alpha_{\ell}\right) {\boldsymbol{S}} \boldsymbol{H}^{(\ell)}+\alpha_{\ell} \boldsymbol{H}^{(0)}\)) is the highlight. By integrating the most 'unsmooth' layer \(\boldsymbol{H}^{(0)}\) during each round of propagation, GCNII circumvents the pitfall described in \textbf{Theorem 1}. In fact, the output feature can still carry information from both the input feature and the graph structure, even as \(K \to \infty\), which is guaranteed by
\textbf{Theorem 4}.
\paragraph{Theorem 4}
A \(K\)-layer GCNII can express a \(K\) order polynomial filter \(\left(\sum_{\ell =0}^K \theta_\ell \boldsymbol{\tilde L}^{\ell}\right)\boldsymbol x\) with arbitrary coefficients.

According the above theorem, by fine-tuning the hyper-parameters \(\alpha_\ell\) and \(\beta_{\ell}\), GCNII can well preserve node features even at high depths. The tuning process would be tedious for a deep network, though.

\section{Future Work}

Although \textbf{Theorem 3} has yielded much theoretical insight into over-smoothing, a tighter bound w.r.t. node features like number of nodes, diameter and sparsity is direly needed. This objective can be better served with comprehensive experiments measuring the effects of those factors on over-smoothing. Also, the interesting properties of residual architectures like GCNII call for further theoretical analysis.

\subsubsection*{Acknowledgments}

This paper is a term project for the Machine Learning course at Peking University, which is taught by Prof. Liwei Wang. The author wishes to express his sincere gratitude to Prof. Wang, who gives him this invaluable opportunity to probe into the exciting world of GCNs.

\section*{References}

\small

[1] Thomas N Kipf\ \& Max Welling. Semi-supervised classification with graph convolutional networks. In {\it Proceedings of the International Conference on Learning Representations}, 2017.

[2] Qimai Li, Zhichao Han,\ \& Xiao-Ming Wu. Deeper insights into graph convolutional networks for semi-supervised learning. In {\it Thirty-Second AAAI Conference on Artificial Intelligence}, 2018a.

[3] Felix Wu, Amauri H. Souza Jr., Tianyi Zhang, Christopher Fifty, Tao Yu,\ \& Kilian Q. Weinberger. Simplifying graph convolutional networks. In {\it ICML}, volume 97 of {\it Proceedings of Machine Learning Research}, pp. 6861-6871. PMLR, 2019.

[4] Hoang NT\ \& Takanori Maehara. Revisiting graph neural networks: All we have is low-pass filters. {\it CoRR}, abs/1905.09550, 2019.

[5] Keyulu Xu, Chengtao Li, Yonglong Tian, Tomohiro Sonobe, Ken-ichi Kawarabayashi,\ \& Stefanie Jegelka. Representation Learning on Graphs with Jumping Knowledge Networks. In {\it Proceedings of the 35th International Conference on Machine Learning}, volume 80, pp. 5453-5462, 2018.

[6] Lingxiao Zhao\ \& Leman Akoglu. Pairnorm: Tackling oversmoothing in GNNs. In {\it International Conference on Learning Representations}, 2020.

[7] Deli Chen, Yankai Lin, Wei Li, Peng Li, Jie Zhou,\ \& Xu Sun.  Measuring and Relieving the Over-smoothing Problem for Graph Neural Networks from the Topological View. In {\it Thirty-Four AAAI Conference on Artificial Intelligence}, 2020.

[8] Kenta Oono\ \& and Taiji Suzuki. Graph neural networks exponentially lose expressive power for node classification. In {\it International Conference on Learning Representations}, 2020.

[9] Miroslav Fiedler. Algebraic connectivity of graphs. {\it Czechoslovak Mathematical Journal}, Vol. 23 ({\bf 1973}), No. 2, 298-305.

[10] Bojan Mohar. The Laplacian spectrum of graphs. {\it Graph Theory, Combinatorics, and Applications}, Vol. 2, Ed. Y. Alavi, G. Chartrand, O. R. Oellermann, A. J. Schwenk, Wiley, 1991, pp.871-898.

[11] Michael S. Cavers. (2010). The normalized Laplacian matrix and general Randic index of graphs. (Doctoral Thesis, University of Regina, Saskatchewan, Canada).

[12] Yu Rong, Wenbing Huang, Tingyang Xu,\ \& Junzhou Huang. DropEdge: Towards deep graph convolutional networks on node classification. In {\it International Conference on Learning Representations}, 2020.

[13] Meng Liu, Hongyang Gao,\ \& Shuiwang Ji. Towards deeper graph neural networks. In {\it ACM SIGKDD Conference on Knowledge Discovery and Data Mining (KDD)}, 2020.

[14] Arman Hasanzadeh, Ehsan Hajiramezanali, Shahin Boluki, Mingyuan Zhou, Nick Duffield, Krishna Narayanan,\ \& Xiaoning Qian. Bayesian graph neural networks with adaptive connection sampling, In {\it Proceedings of the 37th International Conference on Machine Learning}, PMLR 119, 2020.

[15] Cheng Zheng, Bo Zong, Wei Cheng, Dongjin Song, Jingchao Ni,  Wenchao Yu, Haifeng Chen,\ \& Wei Wang, Robust graph representation learning via neural sparsification. In {\it International Conference on Learning Representations}, 2020.

[16] Ming Chen, Zhewei Wei, Zengfeng Huang, Bolin Ding,\ \&  Yaliang Li. Simple and deep graph convolutional networks. In {\it ICML, Proceedings of Machine Learning Research}, pp. 1725-1735. PMLR, 2020.

\end{document}